\documentclass{article}
\usepackage{spconf,amsmath,graphicx}
\usepackage{amssymb}
\usepackage{booktabs}
\usepackage{caption}
\usepackage{xcolor}
\usepackage{subcaption}
\usepackage{indentfirst}
\usepackage{ragged2e}

\title{Fine-grained Spatial-Temporal Perception for Gas Leak Segmentation}
\name{Xinlong Zhao and Shan Du* \vspace{-10pt} \thanks{Thanks to SSHRC for funding support (NFRF-GR024801).}
\thanks{*Corresponding author: shan.du@ubc.ca}}
\address{The University of British Columbia - Okanagan}
\begin{document}
%
\maketitle
\begin{abstract}
Gas leaks pose significant risks to human health and the environment. Despite long-standing concerns, there are limited methods that can efficiently and accurately detect and segment leaks due to their concealed appearance and random shapes. In this paper, we propose a Fine-Grained Spatial-Temporal Perception (FGSTP) algorithm for gas leak segmentation. FGSTP captures critical motion clues across frames and integrates them with refined object features in an end-to-end network. Specifically, we first construct a correlation volume to capture motion information between consecutive frames. Then, the fine-grained perception progressively refines the object-level features using previous outputs. Finally, a decoder is employed to optimize boundary segmentation. Because there is no highly precise labeled dataset for gas leak segmentation, we manually label a gas leak video dataset, GasVid. Experimental results on GasVid demonstrate that our model excels in segmenting non-rigid objects such as gas leaks, generating the most accurate mask compared to other state-of-the-art (SOTA) models.
\end{abstract}

\section{Introduction}
\vspace{-8pt}
Gas leaks are unintended events that can occur due to damaged or faulty connections, corrosion, or improper installation. Whether it is natural gas, methane, or other hazardous gases, their escape into the atmosphere can lead to explosions, fires, and long-term health complications.  Detecting gas leaks is essential for early intervention and mitigation of potential disasters. Traditional leak detection methods involve inspectors manually using instruments near potential leak areas to assess the status of the pipeline. These approaches are inefficient and hazardous to inspectors due to the potentially harmful gas. In contrast, video surveillance systems can provide accurate and real-time monitoring without human labor. Thus, designing a video processing algorithm to recognize leaks by analyzing motion patterns and shape variations is a better option for gas leak segmentation. Such algorithms can process continuous video streams in real time, providing faster and safer detection compared to manual inspections.

Gas leaks possess characteristics distinct from common objects. They are colorless, odorless, and invisible to human senses. Although infrared cameras can visualize leaks \cite{wang2022videogasnet}, distinguishing them often remains challenging due to their non-rigid shape and blurry texture. In many cases, segmentation is only possible when the leak is in motion. For example, Lu et al. \cite{lu2021effective} proposed Gaussian-based background modeling for adaptive gas leak segmentation. However, this method faces significant limitations in industrial scenarios due to the irregular morphology of gas and the variability of the background. Common motion extraction method based on background subtraction and optical flow \cite{yang2021mg} fails due to the variable shape of the object and severe noise disturbances. 

Recently, deep learning techniques have shown robustness in object segmentation \cite{zou2023object}. Fan et al. \cite{fan2021sinetv2} first located coarse concealed objects and then refined them in the decoder. Wang et al. \cite{wang2023deep}, based on YOLOv7 \cite{wang2023yolov7}, demonstrated the capabilities in detecting visible non-rigid objects like dust. For those objects in RGB images, methods relying solely on spatial information may be effective. However, segmenting blurry objects in infrared cameras becomes challenging if models neglect motion information. Yan et al. \cite{yan2019rcrnet} incorporated optical flow to learn motion features, and Yang et al. \cite{yang2021mg} employed unsupervised learning to generate pixel-level motion labels in CNNs. However, both approaches are susceptible to noise interference. Memory-based segmentation methods \cite{bekuzarov2023xmempp, Zhou2024rmem,cheng2024cutie} struggle with the variety of gas leak shapes. Moreover, redundant memory significantly impacts segmentation performance. Cheng et al. \cite{cheng2022sltnet} integrated boundary optimization and optical flow for motion analysis, showing potential for gas leak segmentation. However, it struggles to distinguish objects that are similar to leaks in texture, such as clouds. To the best of our knowledge, few methods can effectively segment blurry and non-rigid objects, like gas leaks, in infrared videos.

\begin{figure*}
    \centering
    \includegraphics[width=1\linewidth]{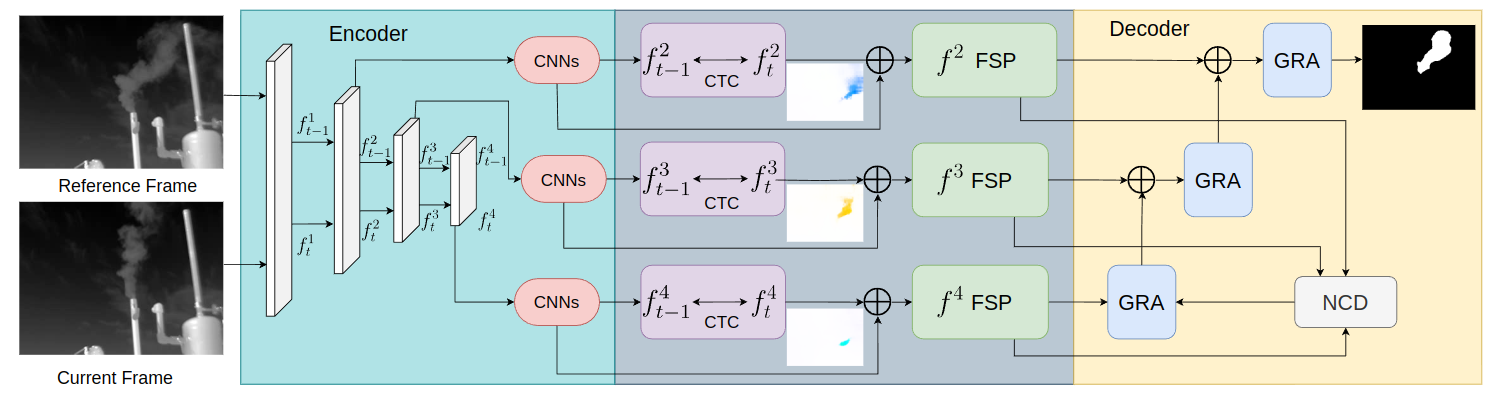}
    \caption{The architecture of FGSTP. The model processes every current frame along with an adjacent frame each time. Each current frame needs to be processed twice with two adjacent frames. The encoder denoises the input frame and extracts the multi-scale features, while the decoder optimizes boundaries through the GRA and NCD modules.  $f_{i}^j$ denotes the feature of the ith frame and jth scale. The Consecutive Temporal Correlation (CTC) block captures motion information. $f^j$ denotes the CTC output in the jth scale. The Fine-grained Spatial Perception (FSP) module refines spatial features.}
    \label{fgstp}
    \vspace{-18pt}
\end{figure*}

In this paper, we address these challenges by proposing a fine-grained spatial-temporal network for non-rigid, blurry object segmentation in infrared video. We adopt a multi-scale transformer-based \cite{wang2022pvt} encoder to mitigate video noise. To extract representative motion features of gas leaks, we refine relevant motion clues across multiple scales using a proposed Consecutive Temporal Correlation module (CTC). We also propose the Fine-grained Spatial Perception module (FSP) to capture spatial features by combining the motion features with encoder features and distilling informative semantic features of leaks. Besides, boundary determination is critical yet challenging for polymorphic objects, as their lack of clear boundaries makes them blend with the background. Our decoder can optimize boundaries from coarse prediction to an accurate mask. We conduct experiments on the GasVid \cite{wang2022videogasnet} dataset, for which we manually annotate gas leak regions frame by frame to ensure the accurate representation of leak area and effective model training. In summary, our contributions include: \textbf{(1)} We propose an end-to-end architecture that is specifically to segment non-rigid and blurry objects in infrared videos; \textbf{(2)} We design CTC and FSP modules to obtain key motion clues and perceive shapes or boundary features from encoder output. 

The paper is organized as follows: Section 2 introduces our FGSTP framework. Section 3 presents experimental results on the GasVid dataset. Section 4 concludes our work.

\vspace{-8pt}
\section{Method}
\vspace{-10pt}
\subsection{Framework Overview}
\vspace{-5pt}
Fig.\ref{fgstp} shows the architecture of FGSTP. The model consists of four components: (1) Transformer-based encoder denoises and obtains multi-scale features; (2) Consecutive Temporal Correlation (CTC) captures key motion; (3) Fine-grained Spatial Perception (FSP) refines spatial features through the residual connection between encoder features and CTC outputs; (4) The decoder optimizes predicted mask boundaries. FGSTP takes the inference frame with two adjacent frames as input to identify the location and shape of the target object. The model output is a binary mask of the inference frame.

FGSTP is a one-stage model and requires no additional auxiliary inputs. It is designed for segmenting blurry non-rigid objects like gas leaks. The shape of leaks changes  significantly in the short term. Our research finds that the two adjacent frames contain sufficient information for the inference frame in this kind of video object segmentation. Therefore our model calculates 3 frames (two adjacent and one inference frame) each time. This operation not only reduces redundant computations but also helps FGSTP capture inter-frame differences more prominently.
\vspace{-8pt}
\subsection{FGSTP Architecture}
\vspace{-5pt}
\textbf{(1) Encoder.} The Pyramid Vision Transformer (PVT) \cite{wang2022pvt} is the model backbone to extract the global features from three input frames. The encoder has four layers, which generate global features at different scales. The size of features in each layer is $C \times H/2^{i+1} \times W/2^{i+1}$, where $i \in \{1,2,3,4\}$, $H,W,C$ are height, width and channels. Previous work finds that the last three layers reserve more high-level semantic features with reduced dimensions \cite{cheng2022sltnet}. Therefore we put the $2nd,3rd,$ and $4th$ layers features into the multi-scale network to extract local features $f_t^i$ ($i \in \{2,3,4\}$).

\begin{figure}
    \includegraphics[width=0.95\linewidth]{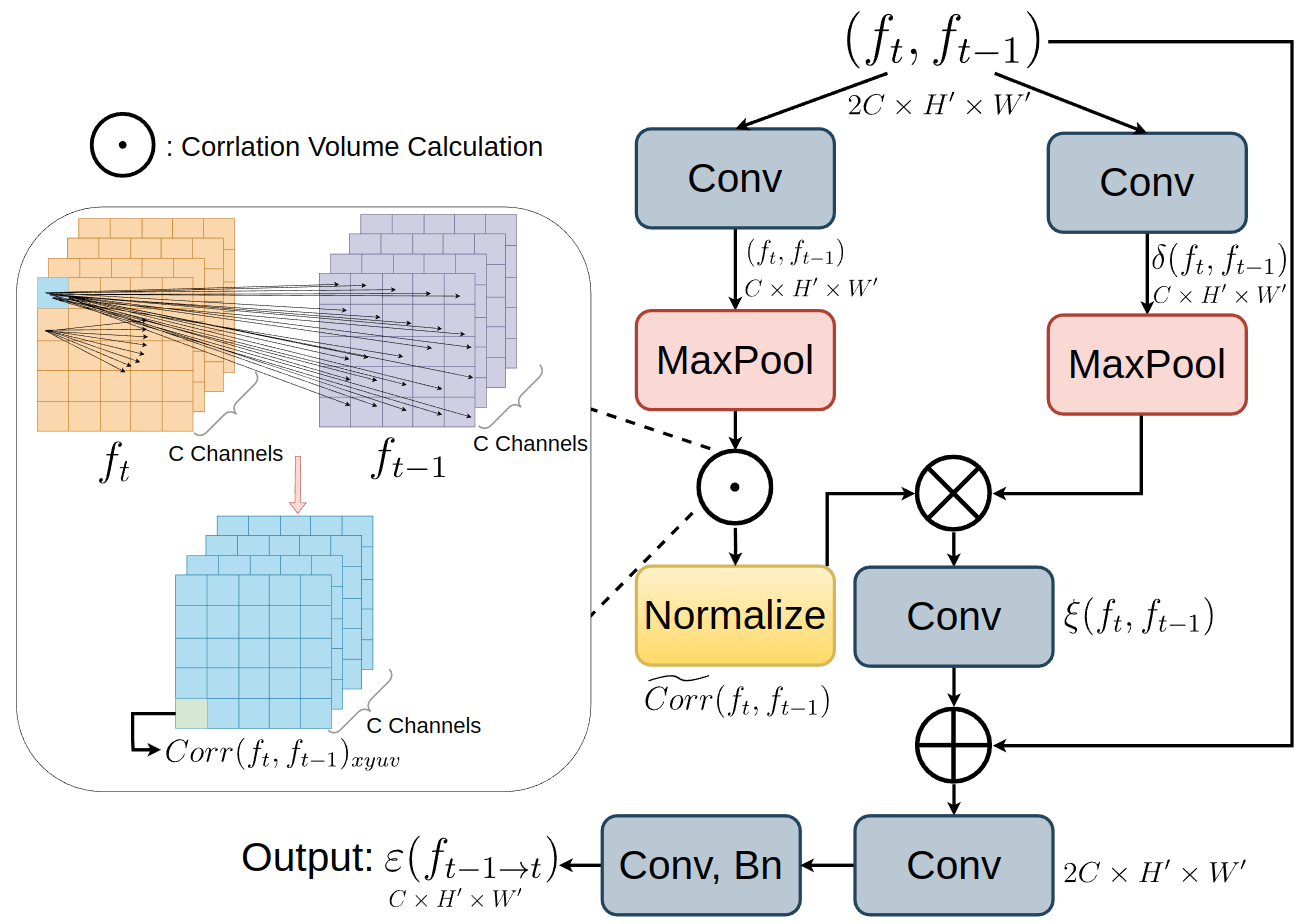}
    \caption{The structure of CTC. It computes 4D correlation volume $Corr(f_t,f_{t-1})$ for motion information and integrates channel-wise semantics for precise motion alignment. }
    \label{ctc}
    \vspace{-15pt}
\end{figure}

\textbf{(2) Consecutive Temporal Correlation.} The Consecutive Temporal Correlation (CTC) is to capture motion clues from consecutive frames. Different from reference \cite{yang2021mg} which needs the optical flow results as the network's input, CTC is an embedded component in FGSTP. Inspired by \cite{li2021arvo}, in each scales ($i \in \{2,3,4\}$), CTC computes temporal correspondence through the feature consistency between two consecutive frames. The structure of CTC is in Fig. \ref{ctc}. Given the features of two frames from the encoder, $(f_{t}, f_{t-1}) \in R^{2C \times H^{\prime} \times W^{\prime}}$, the 4D correlation volume $Corr(f_{t},f_{t-1}) \in R^{H^{\prime} \times W^{\prime} \times H^\prime \times W^\prime}$ can be calculated as follows: 
\vspace{-8pt}

\begin{align}
Corr{(f_{t},f_{t-1})}_{xyuv} = \exp \left(\sum\limits_c {(f_t)}_{xyc} \cdot {(f_{t-1})}_{uvc}\right)
\end{align}
where c is the channel index. When all features similarities are calculated between consecutive frames, the most related part between two frames can be found. Then normalization is applied to correlation volume $Corr{(f_{t}, f_{t-1})}_{xyuv}$ in the latter two dimensions $(u,v)$ to obtain the corresponding part between reference frame and current frame. The normalization is as follows:
\vspace{-3pt}
\begin{align}
\widetilde{Corr}(f_{t},f_{t-1})_{xyuv} = \frac{Corr(f_{t},f_{t-1})_{xyuv}} {\sum_{u}\sum_{v} Corr(f_{t},f_{t-1})_{xyuv}}
\end{align}
The normalized correlation volume is calculated by each channel between features of adjacent frames. It should be merged with the channel features $\delta(f_{t}, f_{t-1})$ to obtain the final temporal correlation. Therefore the channel features multiply normalized correlation and undergo series convolution layers and residual connections to get the final output $\varepsilon(f_{t-1 \rightarrow t})$. The integration between channel-wise features and normalized correlation volume is as follows:
\vspace{-5pt}
\begin{align}
\xi(f_{t-1 \rightarrow t}) = \delta(f_{t},f_{t-1}) \cdot \widetilde{Corr}(f_{t},f_{t-1})
\end{align}

\begin{figure}
    \includegraphics[width=0.95\linewidth]{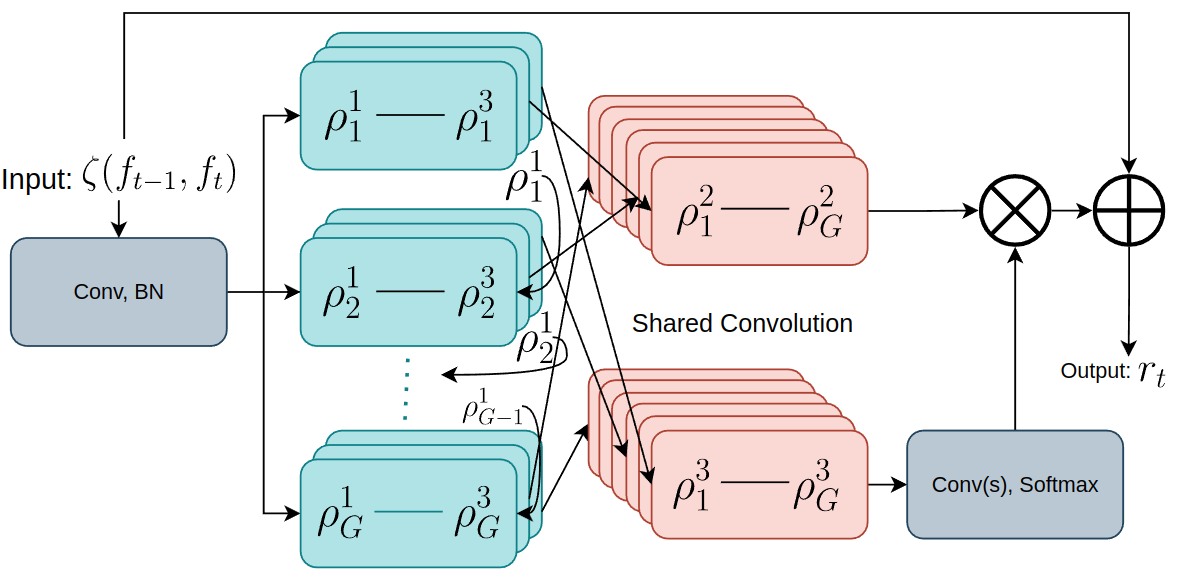}
    \caption{The structure of FSP. It enhances motion features from CTC by residual connection with encoder and calculates semantic representation.}
    \label{fsp}
    \vspace{-15pt}
\end{figure}

\textbf{(3) Fine-grained Spatial Perception.} The output of CTC contains only motion information. Relying solely on motion information is insufficient for gas leak segmentation, as it overlooks valuable details like shape and texture. To address this issue, inspired by \cite{pang2024zoomnext}, we propose Fine-grained Spatial Perception (FSP) to capture spatial features. The input of FSP $\zeta(f_{t-1},f_t)$ combines the CTC ($\varepsilon(f_{t-1\rightarrow t}^i) \in R^{C \times H^\prime \times W^\prime}$) and multi-scale features $f_{t}^i \in R^{C \times H^\prime \times W^\prime}, i \in \{2,3,4\}$ in encoder through residual connection. Fig. \ref{fsp} is the structure of FSP.

 In FSP, firstly the channel number of $\zeta(f_{t-1},f_t)$ is increased and then divided into $G$ groups along the channel dimension, where the features in each group are further split into three sets. The first set ($\rho_j^1, j \in (1...G)$) in each group concatenates with the next group. Then the features of the latter two sets in each group are rearranged into two new groups ($\rho_1^2 \rightarrow \rho_G^2$ and $\rho_1^3 \rightarrow \rho_G^3$) by convolution and split operations according to their position in the original groups. Finally, the two sets of group-mixed features ($\rho^2$ and $\rho^3$) are integrated in a convolution block and produce the output $r_{t}$. The mixing operation extracts critical clues across channels for robust feature representation. The feature group iteration integrates different channel features with partial parameter sharing. These FSP processes improve the ability to capture rich clues, enhancing object perception and prediction refinement.
\vspace{3pt}

\textbf{(4) Decoder.} The decoder is designed to identify the boundaries of non-rigid objects. The Neighbor Connection Decoder (NCD) \cite{fan2021sinetv2} extracts the coarse mask from different scales of the previous module, FSP. The coarse mask is the approximate location of the target object. Reversing the coarse mask means switching the value of foreground and background. Then the coarse mask is interpolated by reversed masks in the Group Reversal Attention (GRA) \cite{fan2021sinetv2} module through concatenation and convolution. 
\vspace{-5pt}
\begin{align}
    q^k = \text{concat}(p_{i}^k : r^k) * W_{GRA}
\end{align}
where $q^k$ is the integration masks of each scale in frame $k$, $p_{i}^k$ is one of the original coarse mask, $r^k$ is the reversed mask, and $W_{GRA}$ is the convolutional kernel of GRA. In this way, only the boundary area is emphasized in the coarse mask because the values of the foreground and background cancel each other out. More details can be seen in \cite{fan2021sinetv2}. Finally we combine the output of GRA and FSP to get the results of gas leak segmentation.
\vspace{-10pt}

\subsection{Loss Function}
\vspace{-5pt}
Like \cite{cheng2022sltnet}, our loss function combines the weighted cross-entropy loss $L_{ce}^{w}$ and intersection-over-union (IoU) loss $L_{iou}^{w}$ to achieve robust optimization. These components balance the accuracy and structural alignment of prediction. The loss function is below:
\vspace{-8pt}
\begin{align}
L_{hybrid} = L_{ce}^{w} + L_{iou}^{w}
\end{align}
The weighted cross-entropy loss, $L_{ce}^{w}$, ensures that the model focuses more on difficult regions by assigning weights to those regions. IoU loss, $L_{iou}^{w}$, complements the cross-entropy loss by emphasizing the overall predicted shape and size consistency with ground truth. 

\vspace{-8pt}
\section{Experiments}
\vspace{-10pt}
\subsection{Dataset}
\vspace{-5pt}
Our experiments use the public dataset, GasVid \cite{wang2022videogasnet}. This dataset is a gas leak video set containing 31 videos, each with a duration of 24 minutes. More details can be found in \cite{wang2022videogasnet}. However, the dataset does not include annotations, so we manually label the videos. In this dataset, some videos are very similar to others, and some are too difficult for humans to distinguish gas leaks. We remove those unused videos and divided the remaining videos into the following categories: close (camera) distance with clear background (no cloud, easy to segment); medium distance with clear background (medium to segment); close distance with complex background (hard to segment); long camera distance with clear background (hard); long camera distance with complex background (hard). For each category, we assign one video for training and another one for testing. This setup allows the model to learn all video types during training and effectively evaluate leak segmentation performance across diverse scenarios. We totally select 14 videos. To accelerate training, each video is divided into many segments that each of them contains 150 frames. We also remove the repeated segments in each video to reduce imbalance. The training set consists of 9 videos with 120 segments, while the testing set has 5 videos with 67 segments totally. Our results on the GasVid dataset are presented in Section 3.3.
\vspace{-8pt}
\subsection{Metrics and Configurations}
\vspace{-5pt}
Our metrics include the following indexes: \textbf{S-measure} ($S_{\alpha}$): proposed in \cite{fan2017smeasure}, combines region-aware and object-aware evaluations to measure structural similarity. \textbf{Weighted F-measure} ($F_{\beta}^{\omega}$) \cite{margolin2014fmeasure}: combines precision and recall, weighted by a balancing factor, to provide a comprehensive evaluation. \textbf{Mean Absolute Error} (MAE): denoted as $M$, quantifies the pixel-level accuracy by comparing the predicted masks with ground truth. \textbf{Enhanced-Alignment Measure} ($E_{\phi}$)\cite{fan2018ephi}: assesses pixel-level correspondence alongside image-level statistics, making it well-suited for evaluating both global and localized accuracy in non-rigid object segmentation. \textbf{Mean Intersection over Union (mIoU)}: calculates the average overlap between prediction and ground truth. \textbf{Mean Dice}: computes the average similarity coefficient, quantifying the overlap between prediction and ground truth, emphasizing segmentation accuracy.

Unlike normal object segmentation methods, we don't use any pre-trained backbone or model due to the specific characteristics of our dataset. Pre-trained models on RGB datasets may negatively impact our experiments. For a fair comparison, we train our model and benchmark models from scratch. The input frames are resized to 352$\times$352. We train for 60 epochs with learning rate $1 \times 10^{-4}$. All models are trained on an NVIDIA RTX 4090 GPU.

\vspace{-8pt}
\subsection{Results}
\vspace{-5pt}
Our experimental results are presented in Table \ref{quantitative}. The green values represent the best scores for each metric, while the red values indicate the second-highest scores. Our method outperforms all benchmark models. Even when compared with the second-best method, SLT-Net \cite{cheng2022sltnet}, our method achieves superior performance on most metrics. Moreover, other methods show significant disparities compared to our model.
\vspace{-5pt}
\begin{table}[ht]
    \centering
    \captionsetup{skip=5pt}
    \caption{Quantitative comparison of various benchmarks on GasVid dataset.}
    \label{quantitative}
    \scalebox{0.8}{ 
    \begin{tabular}{ccccccc}
    \toprule
    Models & \textbf{$S_{\alpha}\uparrow$} & \textbf{$F_{\beta}^{\omega}\uparrow$} & \textbf{$M\downarrow$} & \textbf{ $E_{\phi}\uparrow$} & $mIOU\uparrow$ &  $mDice\uparrow$ \\ 
    \midrule
    MG \cite{yang2021mg} & - & - & - & - & - & - \\
    SINet-V2 \cite{fan2021sinetv2} &0.684  & 0.472 & \textcolor{green}{0.022} & 0.743 & 0.361 & 0.465 \\
    XMem++ \cite{bekuzarov2023xmempp} & 0.675  & 0.424  & 0.036 & 0.675 & 0.361 & 0.459 \\ 
    RMem \cite{Zhou2024rmem} & 0.685  & 0.431  &\textcolor{red}{0.023} & 0.711 & 0.361 & 0.451 \\
    Zoomnext \cite{pang2024zoomnext} & 0.691 & 0.450 & 0.025 & 0.729 & 0.378 & 0.472 \\
    SLT-Net \cite{cheng2022sltnet} & \textcolor{red}{0.702}  & \textcolor{red}{0.500}  & 0.025 & \textcolor{green}{0.806} & \textcolor{red}{0.392} &\textcolor{red}{0.505} \\ 
    \textbf{FGSTP (Ours)} & \textcolor{green}{0.705}  & \textcolor{green}{0.507} & \textcolor{green}{0.022} & \textcolor{red}{0.797} & \textcolor{green}{0.399} & \textcolor{green}{0.509} \\ 
    \bottomrule
    \end{tabular}
    }
    \vspace{-10pt}
\end{table}

\begin{figure*}[ht]
    \begin{subfigure}[c]{1\textwidth}
        \begin{minipage}{0.015\textwidth}
            \centering 
            \rotatebox{90}{1467}
        \end{minipage}
        \begin{minipage}{0.9\textwidth}
            \centering
            \includegraphics[scale=0.22]{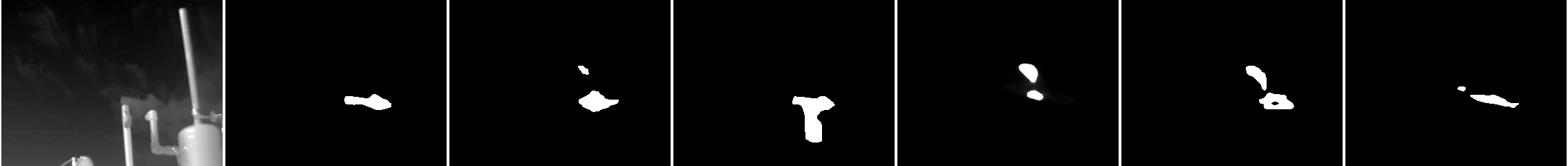}
        \end{minipage}
    \end{subfigure}
    \vspace{0.05cm}
    \begin{subfigure}[c]{1\textwidth}
        \begin{minipage}{0.015\textwidth}
            \centering
            \rotatebox{90}{1476} 
        \end{minipage}
        \begin{minipage}{0.9\textwidth}
            \centering
            \includegraphics[scale=0.22]{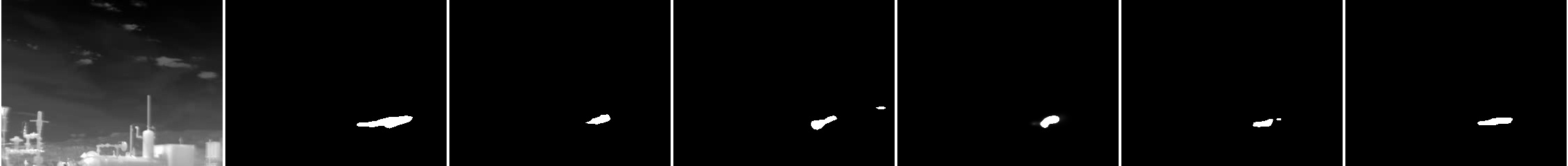}
        \end{minipage}
    \end{subfigure}
    \vspace{0.05cm}
    \begin{subfigure}[c]{1\textwidth}
        \begin{minipage}{0.015\textwidth}
            \centering
            \rotatebox{90}{2559}
        \end{minipage}
        \begin{minipage}{0.9\textwidth}
            \centering
            \includegraphics[scale=0.22]{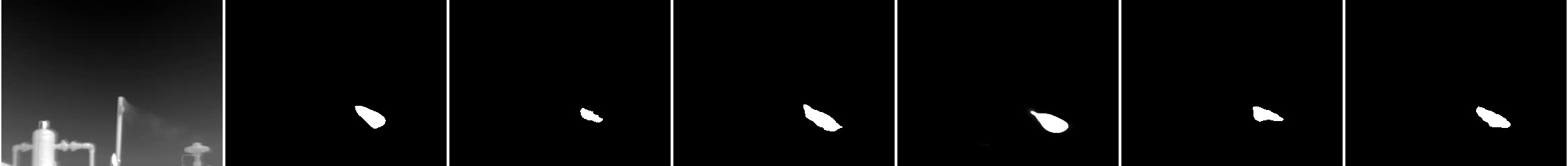}
        \end{minipage}
    \end{subfigure}
    \vspace{0.05cm}
    \begin{subfigure}[c]{1\textwidth}
        \begin{minipage}{0.015\textwidth}
            \centering
            \rotatebox{90}{2563} 
        \end{minipage}
        \begin{minipage}{0.9\textwidth}
            \centering
            \includegraphics[scale=0.22]{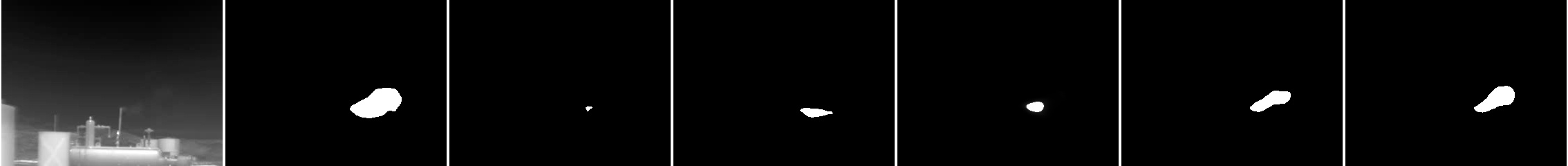}
        \end{minipage}
    \end{subfigure}    
    \vspace{0.05cm}
    \begin{subfigure}[c]{1\textwidth}
        \begin{minipage}{0.015\textwidth}
            \centering
            \rotatebox{90}{2566}
        \end{minipage}
        \begin{minipage}{0.9\textwidth}
            \centering
            \includegraphics[scale=0.22]{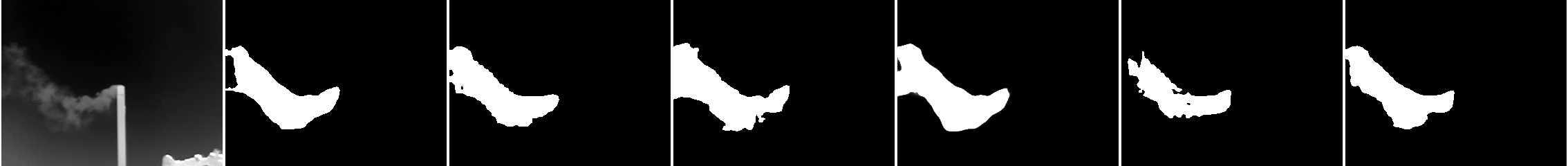}
        \end{minipage}
    \end{subfigure}
    \vspace{0.05cm}
    \begin{tabular}{p{0.5cm} p{2.3cm} p{1.6cm} p{2.1cm} p{2.0cm} p{2.2cm} p{2.0cm} p{2.0cm}}
        & \small Image 
        & \small GT 
        & \small SINet-V2 \cite{fan2021sinetv2} 
        & \small RMem \cite{Zhou2024rmem} 
        & \small Zoomnext \cite{pang2024zoomnext} 
        & \small SLT-Net \cite{cheng2022sltnet} 
        & \small FGSTP (Ours) \\
    \end{tabular}
    \vspace{-10pt}
    \caption{Visualization results on GasVid Dataset. Our model prediction is the most accurate in different situations, $i.e.,$ close (camera) distance with complex background (cloud or birds interference, 1467), long distance with complex background (1476), medium distance with clear background (2559), long distance with clear background (2563), close distance with clear background (2566)}
    \label{visualization}
    \vspace{-10pt}
\end{figure*}

Table \ref{quantitative} presents the performance comparison across different models. MG \cite{yang2021mg} fails on the GasVid dataset, producing either incomplete or missing masks due to its reliance solely on optical flow without supervised learning. This approach is ineffective for segmenting blurry and non-rigid objects, like gas leaks. SINet-V2 \cite{fan2021sinetv2}, an image-based method, processes each video frame independently and ignores motion information, which significantly affects its performance in segmenting leaks. As a result, it achieves the lowest mIoU (0.361). XMem++ and RMem \cite{bekuzarov2023xmempp, Zhou2024rmem} are memory-based models that require the first-frame mask as input to guide segmentation. Although effective for regular objects, they struggle with non-rigid and blurry objects, leading to lower weighted F-measures (0.424 and 0.431) compared to SINet-V2 (0.472). ZoomNext \cite{pang2024zoomnext} leverages strong spatial feature extraction to perform better than image-based and memory-based models, reaching 0.378 mIoU. However, its limited motion extraction capability hinders its performance for tracking leaks. SLT-Net \cite{cheng2022sltnet} enhances SINet-V2 by incorporating motion information through optical flow, improving performance to 0.392 mIoU. However, it still lacks object-level feature representation, limiting its segmentation capability. In contrast, our model, FGSTP, integrates fine-grained spatial perception with precise motion details, allowing it to effectively find leaks, and achieves the highest performance, 0.399 mIoU.

We also provide visualized results in Fig.\ref{visualization}. Due to MG's failure and space limitations, we select the results of the best five models on all test videos. Our model consistently produces the most accurate masks across various scenarios. For Video-1467 (first row), the background contains clouds and birds, with the clouds having appearances similar to gas leaks. In these challenging cases, only our model correctly segments the leak. Video-1476 is the most difficult sample because of its long camera distance and cloud-covered background. All models, except ours, generate masks with incorrect shapes and locations. Video-2559 also involves a long camera distance but lacks clouds. SLT-Net correctly localizes the leak in this video, but unlike our model, it fails to accurately match its shape. Videos 2559 and 2566 are relatively easier and all models successfully segment leaks; however, our model's predictions have more details and ours are the most similar to the ground truth (GT). These results demonstrate that our model not only captures more details in simpler scenarios but also shows more stable and accurate performance in challenging conditions.

\vspace{-15pt}
\subsection{Ablation Studies}
\vspace{-5pt}
We conduct ablation studies on the GasVid dataset to evaluate the effectiveness of the two core modules: CTC and FSP. The results are presented in Table \ref{ablation}. Without both CTC and FSP, the model lacks both spatial and temporal information, resulting in the worst performance with only 0.291 mIoU, even worse than many benchmark models. This is because the decoder struggles to generate accurate masks using only backbone features. Removing FSP while keeping CTC eliminates spatial perception, causing the performance to drop to 0.383 mIoU. Although worse than the full model, it still performs better than without both modules or without CTC. Removing CTC and remaining FSP leads to a more significant performance drop to 0.313 mIoU, indicating that motion features are very important. However, spatial perception also contributes to the model performance. The ablation results demonstrate that the superior performance of our model comes from the integration of CTC and FSP, which maximizes the benefits of both temporal and spatial dimensions.
\vspace{-10pt}
\begin{table}[ht]
    \centering
    \captionsetup{skip=5pt}
    \caption{Ablation studies results of CTC and FSP modules in FGSTP on GasVid dataset}
    \label{ablation}
    \scalebox{0.75}{
    \begin{tabular}{ccc|cccccc}
    \toprule
     Backbone &CTC &FSP  & \textbf{$S_{\alpha}\uparrow$} & \textbf{$F_{\beta}^{\omega}\uparrow$} & \textbf{$M\downarrow$} & \textbf{$E_{\phi}\uparrow$} & $mIOU\uparrow$ & $mDice\uparrow$ \\ 
    \midrule
    \checkmark & & &0.644 &0.419 &0.027 &0.741 &0.291 &0.408  \\
    \checkmark &\checkmark  &  &0.697  &0.488  &0.024  &0.782  &0.383  &0.488  \\ 
    \checkmark &  &\checkmark &0.655  &0.430  &0.027  &0.761  &0.313 & 0.428 \\
    \checkmark &\checkmark &\checkmark & \textbf{0.705}  & \textbf{0.507} & \textbf{0.022} & \textbf{0.797} & \textbf{0.399} & \textbf{0.509} \\ 
    \bottomrule
    \end{tabular}
    }
\end{table}
\renewcommand{\arraystretch}{1}

\section{Conclusion}
\vspace{-5pt}
This paper presents an innovative method for gas leak segmentation that achieves better accuracy and adaptability across diverse scenarios. Our approach incorporates an encoder to reduce noise, a temporal module to refine motion clues, and a spatial perception to effectively distinguish the target object. The experimental results demonstrate that our model outperforms the SOTA benchmarks. Additionally, ablation studies confirm the critical role of our two core modules for robust performance. Finally, visualized results highlight that our model not only captures finer details in easier cases but also handles various challenging situations effectively.

\vfill\pagebreak

\bibliographystyle{IEEEbib}
\bibliography{strings,refs}

\end{document}